\pdfoutput=1

\documentclass[11pt]{article}

\usepackage{acl}

\usepackage{times}
\usepackage{latexsym}

\usepackage[T1]{fontenc}

\usepackage[utf8]{inputenc}

\usepackage{microtype}

\usepackage{inconsolata}

\usepackage{graphicx}
\usepackage{amsmath}
\usepackage{tabularray}

\UseTblrLibrary{diagbox}

\title{LLaSA: \textbf{L}arge \textbf{L}anguage and \textbf{S}tructured Data \textbf{A}ssistant}

\author{
Yao Xu\textsuperscript{1,2},\, 
Shizhu He\textsuperscript{1,2},\, 
Jiabei Chen\textsuperscript{1,2},\,
Xiangrong Zeng\textsuperscript{3},\,
\\
\textbf{Bingning Wang}\textsuperscript{3},\,
\textbf{Jun Zhao}\textsuperscript{1,2},\,
\textbf{Kang Liu}\textsuperscript{1,2,4\thanks{ \;Corresponding Author}} \\
\textsuperscript{1} The Laboratory of Cognition and Decision Intelligence for Complex Systems, \\ Institute of Automation, Chinese Academy of Sciences, Beijing, China\\ 
\textsuperscript{2} School of Artificial Intelligence, University of Chinese Academy of Sciences, Beijing, China \\ 
\textsuperscript{3} Baichuan Inc, Beijing, China  \\ 
\textsuperscript{4} Shanghai Artificial Intelligence Laboratory, Shanghai, China  \\ 
\{yao.xu, shizhu.he, jzhao, kliu\}@nlpr.ia.ac.cn, chenjiabei2024@ia.ac.cn\\
}

\begin{document}
\maketitle
\begin{abstract}
Structured data, such as tables, graphs, and databases, play a critical role in plentiful NLP tasks such as question answering and dialogue system. 
Recently, inspired by Vision-Language Models, Graph Neutral Networks (GNNs) have been introduced as an additional modality into the input of Large Language Models (LLMs) to improve their performance on Structured Knowledge Grounding (SKG) tasks.
However, those GNN-enhanced LLMs have the following limitations: 
(1) They employ diverse GNNs to model varying types of structured data, rendering them unable to uniformly process various forms of structured data.
(2) The pretraining of GNNs is coupled with specific LLMs, which prevents GNNs from fully aligning with the textual space and limits their adaptability to other LLMs.
To address these issues, we propose \textbf{L}arge \textbf{L}anguage and \textbf{S}tructured Data \textbf{A}ssistant (LLaSA), a general framework for enhancing LLMs' ability to handle structured data. Specifically, we represent various types of structured data in a unified hypergraph format, and use self-supervised learning to pretrain a hypergraph encoder, and a G-Former compressing encoded hypergraph representations with cross-attention.
The compressed hypergraph representations are appended to the serialized inputs during training and inference stages of LLMs.
Experimental results on multiple SKG tasks show that our pretrained hypergraph encoder can adapt to various LLMs and enhance their ability to process different types of structured data. Besides, LLaSA, with LoRA fine-tuning, outperforms previous SOTA method using full parameters tuning.
\end{abstract}

\section{Introduction}

\begin{figure}[t]
  \centering
  \includegraphics[width=0.4\textwidth]{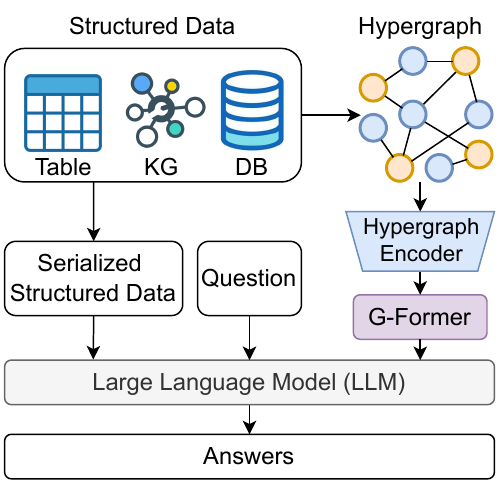} 
  \caption{Overview of LLaSA, which can handle various types of structured data by transforming them into a unified format and encoding them with a universal encoder. The serialized structured data and the graph representations are then used as input to the LLM.}
  \label{fig:llasa_overview}
\end{figure}

Structured data, such as tables, knowledge graphs, and databases, is prevalent in real-world applications and plays a crucial role in fields like finance, healthcare, and data analytics. Therefore, Structured Knowledge Grounding (SKG) ~\citep{xie_unifiedskg_2022} has attracted significant research interest and has been widely studied. SKG tasks, such as question answering~\citep{pasupat2015wtq, nan2022fetaqa, talmor2018cwq}, summarization~\citep{nan2020dart,parikh2020totto}, fact verification~\citep{chen2019tabfact}, utilizing corresponding structured data as input and produce different outputs depending on the task types.

In recent years, with the rapid development of Large Language Models (LLMs)~\citep{eval_chatgpt, zhao2023survey},  researchers have shifted their focus from building task-specific models for different tasks~\citep{xie_unifiedskg_2022} to developing a generalist model capable of handling a variety of SKG tasks~\citep{zhuang_structlm_2024,zhang_tablellama_2024}. These approaches that leverage LLMs for SKG tasks commonly serialize structured data (e.g., representing tables in markdown format) as pure textual input to the LLMs. 
However, this method can lead to the partial loss of structured information, as all these LLMs are decoder-only Transformer models \citep{vaswani_attention_2017} (e.g., in the table data, cells from the same column or rows in the original table may become distant from each other after linear serialization).

\begin{figure}[t]
  \centering
  \includegraphics[width=0.48\textwidth]{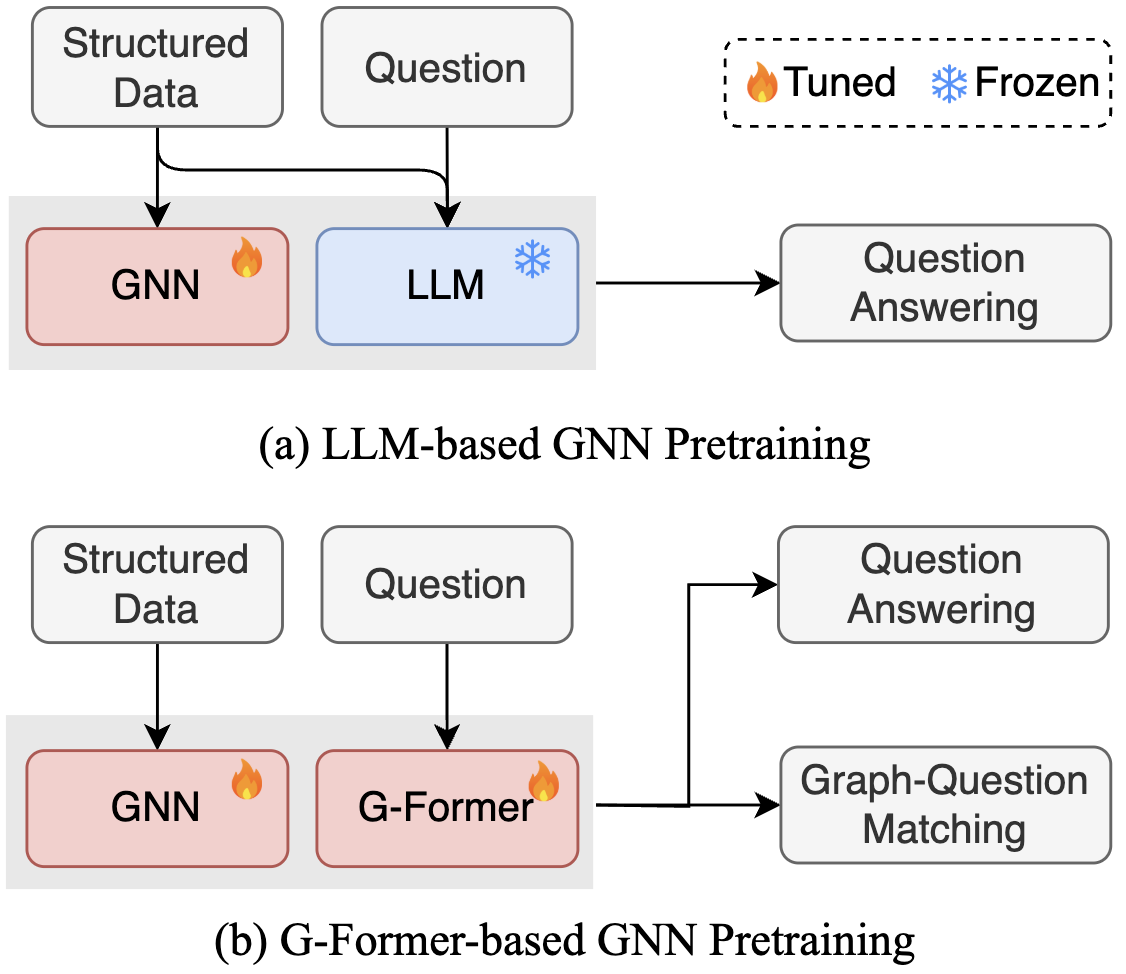}
  \caption{Comparison between LLM-based and the proposed G-Former-based GNN pretraining strategies.}
  \label{fig:pretraining_strategies}
\end{figure}

Recently, to enhance the utilization of large language models in the visual domain, researchers have crafted Vision-Language Models (VLM) \cite{zhang2024vlm} that transform image data into discrete language tokens via a learnable interface.
Inspired by the success of VLM, another line of research introduces GNNs as an additional modality into the input of LLMs. For example, G-Retriever \cite{he_g-retriever_2024} combines GNNs encoding knowledge graphs with LLMs, enhancing the graph-based question-answering abilities of the LLMs. HGT \cite{jin_hgt_2024} propose a heterogeneous graph enhanced large language model for table-based question answering.
However, \textbf{they employ diverse networks to model varying types of structured data, rendering them unable to uniformly process various forms of structured data}, for instance, G-Retriever and HGT can only handle graphs and tables, respectively.
Besides, the pretraining of GNNs is coupled with specific LLMs in these methods, for instance, HGT pretrains a GNN based on a frozen LLM by self-supervised learning, as shown in Figure \ref{fig:pretraining_strategies} (a). \textbf{This prevents the GNN from fully aligning with the textual embedding space because the serialized table is already included as input during the pretraining process, making the GNN unnecessary in this situation.} As a result, it is unclear whether the GNN effectively encodes the table data as expected during pretraining, and the adaptability of this GNN to other LLMs also remains a question.


Aiming to address these drawbacks, we introduce \textbf{L}arge \textbf{L}anguage and \textbf{S}tructured Data \textbf{A}ssistant (LLaSA) for SKG tasks. 
Specifically, we first model various forms of structured data, such as tables and knowledge graphs, uniformly as hypergraphs \citep{chen_hytrel_2023}, enabling the use of a unified GNN for encoding. Specifically, We treat the cells in a table as nodes, with rows and columns as hyperedges, and for graphs, we treat entities as nodes and relationships as hyperedges.
We then pretrain a GNN and a G-Former (a cross attention model similar to Q-Former \cite{li_blip-2_2023}, but it extracts features from GNN) with self-supervised learning which includes question answering and Graph-Text Matching, as illustrated in Figure \ref{fig:pretraining_strategies} (b). This pretraining approach not only aligns the GNN with the text more effectively but also avoids coupling with a specific LLM, making it adaptable to various LLMs.
During fine-tuning for downstream tasks, we use the G-Former to bridge the modality gap, transforming the encoded structured data into a fixed number of soft tokens that can be understood by LLMs, as shown in Figure \ref{fig:llasa_overview}.

Results on multiple SKG datasets, including table, knowledge graph and database, demonstrate that the proposed LLaSA significantly enhances LLM's ability to handle these structured data. With the frozen LLM, LLaSA Llama-7B achieves an average improvement of 12\% across ten datasets. With LoRA the tuned LLM, it still yields an average improvement of 0.4\%. Besides, LLaSA, with LoRA fine-tuning, outperforms previous SOTA method using full parameters tuning. The codes and data are available at \url{https://github.com/YaooXu/LLaSA}. 

The main contributions of this paper can be summarized as follows:
\begin{enumerate}
    \item We propose LLaSA, a framework that integrates the encoded representations of structured data as an additional modality into the input of LLMs.
    \item We represent various forms of structured data as hypergraphs, enabling unified encoding through a single GNN, and pretrain the GNN and G-Former with self-supervised learning.
    \item Experimental results demonstrate that our pretrained GNN can be adapted to various LLMs, enhancing their ability to handle structured data. Furthermore, ablation studies confirm the importance of both the GNN and the pretraining process.
\end{enumerate}

\section{Related Work}
\subsection{Models for SKG tasks}
SKG data, such as graphs and tables, exhibit heterogeneous data formats, leading to a line of research focuses on modeling these heterogeneous representations during encoding  structured data. For example, 
TaBERT \cite{yin_tabert_2020} introduces vertical self-attention, a self-attention mechanism that processes vertically aligned vectors across different rows. TAPAS \cite{herzig_tapas_2020} captures tabular structure with additional embeddings, such as Column / Row ID, based on BERT’s architecture \cite{devlin_bert_2019}. HyTrel \cite{chen_hytrel_2023} converts a table into a hypergraph to allow the GNN to incorporate row/column permutation invariances. All these methods can also be used in LLaSA, and we use HyTrel as our default hypergraph encoder in this work.

USKG \cite{xie_unifiedskg_2022} is the first work that unifies multiple SKG tasks into a text-to-text format. However, their results show that multi-task finetuning is worse than single-task finetuning on many tasks.
Following USKG, StructLM \cite{zhuang_structlm_2024} finetunes LLMs on multiple SKG tasks and show strong zero-shot generalization capability on unseen SKG tasks. TableLlama \cite{zhang_tablellama_2024} finetunes LLMs with LongLoRA \cite{chen2023longlora} on multiple table-based datasets to build a generalist model. However, these methods all serialize structured data and could lead to the partial loss of structured information.

\subsection{Combine LLMs and GNN}
There are many works that combine LMs and GNNs \cite{malaviya2020commonsense,yasunaga2022deep,zhang2022greaselm,zhao2022learning}. In the era of LLMs, researchers are increasingly focused on how to convert GNN representations into tokens that LLMs can understand, thereby avoiding modifications to the model architecture and minimizing the impact on other capabilities.
LLaGA \cite{chenLLaGALargeLanguage2024} reorganizes graph nodes to structureaware sequences and then mapping these into the token embedding space through a projector.
G-Retriever \cite{he_g-retriever_2024} uses a standard Graph Attention Network (GAT) \cite{velivckovic2017gat} to encode the retrieved graphs and treats graph embeddings as soft prompting, but it doesn't involve pretraining stage. 
GraphGPT \cite{tang_graphgpt_2023} not only appends the representations of graph nodes to the textual input, but also employs self-supervised training to align the encoding of graph structures with the natural language space. Another similar work is HGT \cite{jin_hgt_2024}, which introduces an GNN to encode the heterogeneous graph converted by the corresponding table. 
Both GraphGPT and HGT need to pretrain a GNN or a adapter based on a frozen LLM by self-supervised learning before task-specific instruction tuning. In contrast, our LLaSA pretrain a general GNN and G-Former that are decoupled from the specific LLM, allowing them to be used with any LLM without the need for re-pretraining, which is time-consuming.

\section{Method}

\subsection{Hypergraph Construction}

\begin{figure}[t]
  \centering
  \includegraphics[width=0.45\textwidth]{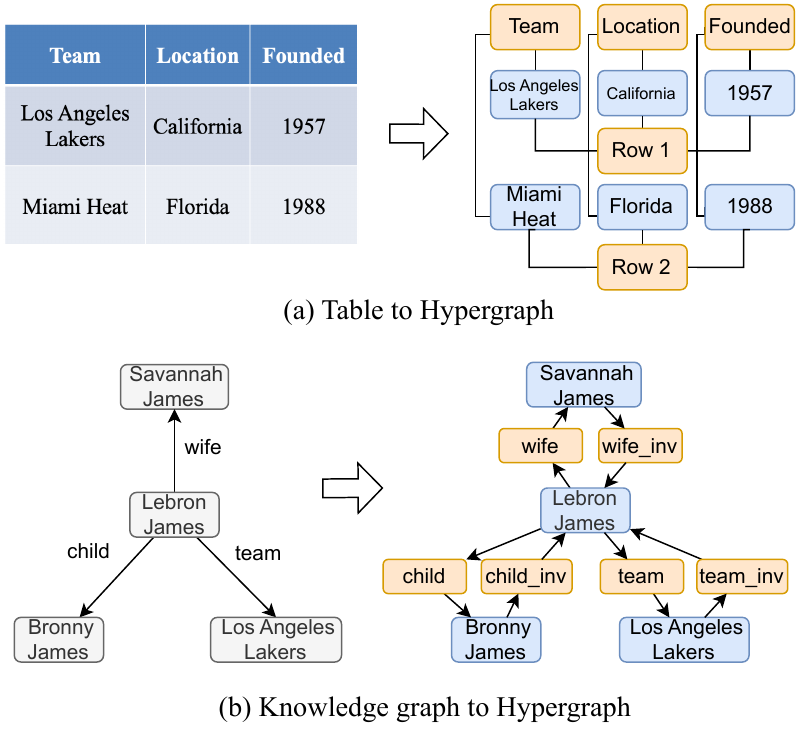} 
  \caption{Examples of converting structured data to a unified hypergraph format, where yellow nodes represent hyperedges. In Figure (a), the arrows are omitted as the edges in the hypergraph are bidirectional.}
  \label{fig:hypergraph}
\end{figure}

\begin{figure*}[t]
  \centering
  \includegraphics[width=\textwidth]{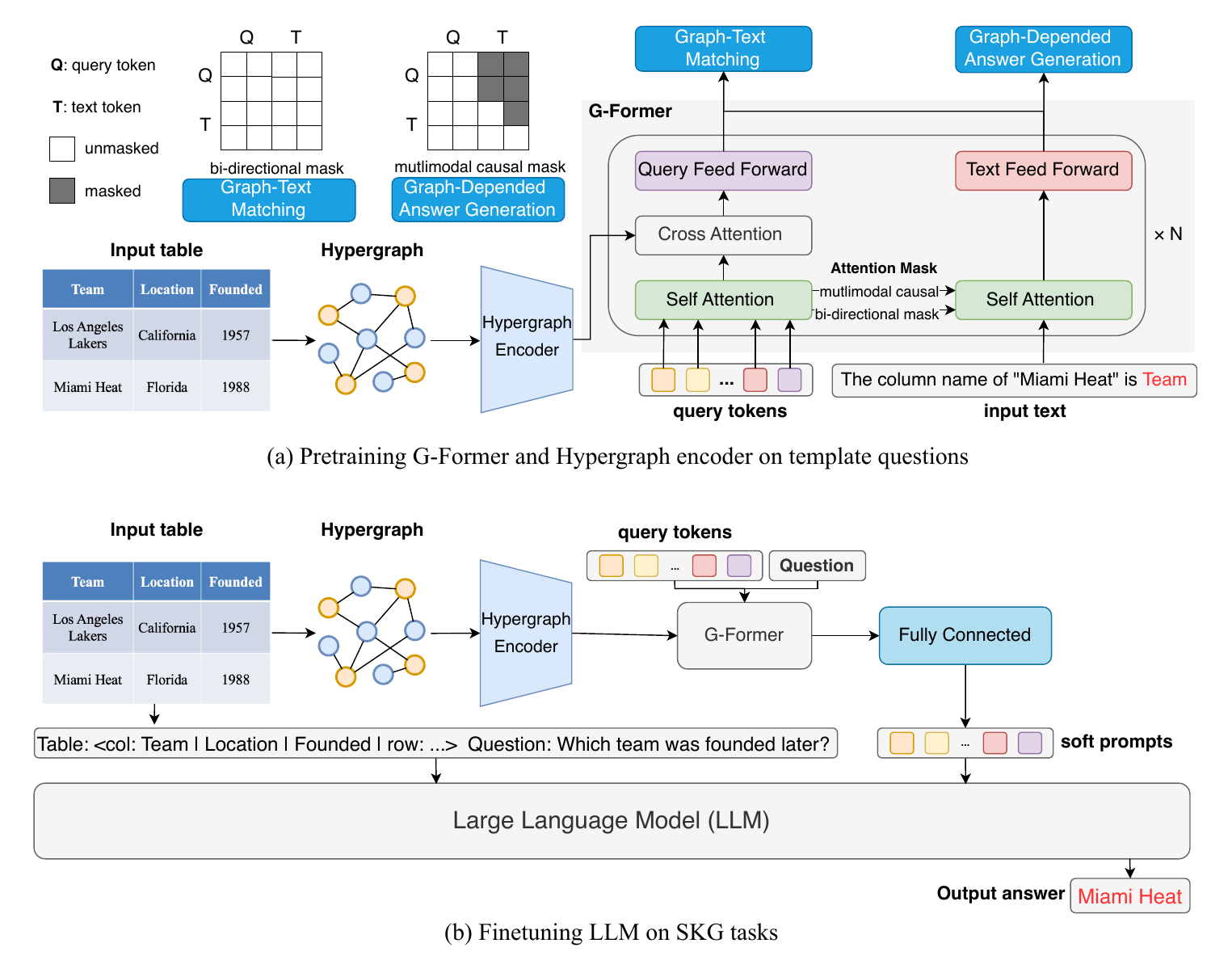} 
  \caption{(a) We employ two pretraining objectives to train the hypergraph encoder and G-Former, with the upper left corner showing the attention masks used for different pretraining tasks. (b) In the LLM finetuning stage, we only use the graph transformer to extract a fixed number of representations of hypergraph, and treat them as soft prompts in LLM's input. }
  \label{fig:method}
\end{figure*}

We represent a hypergraph as $\mathcal{G} = \{\mathcal{V}, \mathcal{E}\}$, where $\mathcal{V}$ and $\mathcal{E}$ denote the set of nodes and hyperedges. A hypergraph can be regarded as a type of bipartite graph, that is, every edge connects a node in $\mathcal{V}$ to one in $\mathcal{E}$.

\noindent \textbf{Table to hypergraph}. We represent a table as $\mathcal{T} = \{\mathcal{H}, \mathcal{R}\}$, where $\mathcal{H}=[h_1, h_2,..., h_n]$ represents $n$ column headers, $\mathcal{R}=[r_1, r_2,..., r_m]$ represents $m$ rows, and each row $r_i=[c_{i1},c_{i2},...,c_{in}]$ has $n$ cells. We treat each cell $c_{ij}$ as node $v_{ij} \in \mathcal{V}$, and each row $r_i$, each column header ${h_j}$ as hyperedges $e_{i}, e_{j} \in \mathcal{E}$. Each node $v_{ij}$ is only connected to its corresponding hyperedges $e_{i}$ and $e_{j}$, as shown in Figure \ref{fig:hypergraph} (a).

\noindent \textbf{Graph to hypergraph}. In this work, our research focuses on Text-Attributed Graphs (TAG), which are graphs enriched with textual information associated with their nodes or edges. We represent a TAG as a set of factual triples, i.e., $G = \{(h,r,t)\}$, where $h, r, t$ are texts, and $h, t$ denote the head and tail entity, $r$ denotes the relation of them. We treat $h$ and $t$ as nodes $v_{h}, v_{t} \in \mathcal{V}$, $r$ as hyperedge $e_{r} \in \mathcal{E}$, and $v_{h}$, $v_{t}$ are connected to $e_{r}$. Besides, to preserve the directional information in the original graph, we create a reverse relation node for each relation node, as shown in Figure \ref{fig:hypergraph} (b) (e.g., the texts of normal/reverse relation node are \textit{Relation: team} and \textit{Reverse Relation: team}, respectively).


\subsection{Model Architecture}

\subsubsection{Hypergraph Encoder}

Following HyTrel \cite{chen_hytrel_2023}, we utilize a HyperTrans, which is a structure-aware transformer module, to encode the hypergraphs. Each layer of HyperTrans contains two attention module: Node2Hyperedge and Hyperedge2Node, and a Hyperedge Fusion module. The initial representation of nodes are obtained by sentence bert \cite{reimers2019sentence_bert}.

The Node2Hyperedge attention module aggregates information to hyperedge $e$ from its neighbor nodes $v \in \mathcal{N}_e$. This process is defined as follows:
\begin{equation}
    \tilde{\mathbf{h}}_e^{l+1} = f_{\mathcal{V} \rightarrow \mathcal{E}} (K_e^l)
\label{eq:1}
\end{equation}
where $f_{\mathcal{V} \rightarrow \mathcal{E}}$ is a attention function, $K_e^l = \{\mathbf{h}_v^l | v \in \mathcal{N}_e\}$ represents the set of representations at layer $l$ of all nodes connected to the hypernode $e$ .

The Hyperedge Fusion module is a Multilayer Perceptron (MLP) that integrates the information collected from both the neighbors of hypernode \( e \) and itself. This process is defined as follows:
\begin{equation}    \mathbf{h}_e^{l+1}=\operatorname{MLP}(\mathbf{h}_e^{l};\tilde{\mathbf{h}}_e^{l+1})
\label{eq:2}
\end{equation}

The Hyperedge2Node attention module then aggregates information to node $v$ from its neighbor hypernodes $e \in \mathcal{N}_v$.
\begin{equation}
    \tilde{\mathbf{h}}_v^{l+1} = f_{\mathcal{E} \rightarrow \mathcal{V}} (K_v^l)
\label{eq:3}
\end{equation}
where $f_{\mathcal{E} \rightarrow \mathcal{V}}$ is another attention function, $K_v^l = \{\mathbf{h}_e^l | e \in \mathcal{N}_v\}$ represents the set of representations at layer $l$ of all hyernodes connected to the node $e$.

The attention function $f$ used in the equation (\ref{eq:1}, \ref{eq:3}) is similar to transformer \cite{vaswani_attention_2017}, the function $f$ is defined  as follows:
\begin{align}
    f_{\mathcal{V} \rightarrow \mathcal{E} \, or \, \mathcal{E} \rightarrow \mathcal{V}} (\mathbf{X}) = \operatorname{LN}(\mathbf{Y} + \operatorname{FFN}(\mathbf{Y})) \\
    \mathbf{Y} = \operatorname{LN}(\omega+\operatorname{SetMHA}(\omega,\mathbf{X}, \mathbf{X}))
\end{align}
where $\mathbf{X}$ is the representations of input nodes or hyperedges. $\mathbf{Y}$ is the intermediate representations. SetMHA is the multi-head set attention mechanism defined as follows (for simplicity, we only consider single-head self-attention here):
\begin{equation}
\small
    \operatorname{SetMHA}(\omega,\mathbf{X}, \mathbf{X})=\operatorname{Softmax}(\omega(\mathbf{X}{\mathbf{W}^K})^T(\mathbf{X}{\mathbf{W}^V}))
\end{equation}
where $\omega$ is a learnable query vector, $\mathbf{W}^K$ and $\mathbf{W}^V$ are the key and value matrices.

In summary, HyperTrans first updates the representation of a hypernode based on the neighboring nodes, and then updates the neighboring nodes using the updated representation of the hypernode.

\subsubsection{G-Former}
To bridge the gap between a hypergraph encoder and text, and to compress the hypergraph node representations into fixed-length tokens, we propose G-Former based on Q-Former \cite{li_blip-2_2023}. 
As demonstrated in Figure \ref{fig:method} (a), our G-Former consists of two transformer sub-modules: (1) A graph transformer which interacts with hypergraph representations. 
(2) A text transformer that encodes and generates text. The graph transformer uses a fixed number of learnable query tokens, which first interact with each other through self-attention, then interact with hypergraph nodes representation through cross-attention.
During the pretraining stage, we use different attention mechanisms based on the specific pretraining tasks to control the interaction between query and text tokens.

\begin{table*}
\tiny
\centering
\begin{tblr}{
  width = \linewidth,
  rows={1.4em, m, rowsep=1pt},
  colspec = {Q[180]Q[54]Q[60]Q[56]Q[60]Q[58]Q[56]Q[50]Q[50]Q[48]Q[48]Q[44]Q[44]Q[48]Q[48]Q[44]},
  row{2} = {c},
  row{3} = {c},
  row{4} = {c},
  row{8} = {c},
  row{12} = {c},
  cell{1}{2} = {c=11}{0.584\linewidth,c},
  cell{1}{13} = {c=4}{0.184\linewidth,c},
  cell{4}{1} = {c=16}{0.916\linewidth},
  cell{5}{2} = {c},
  cell{5}{3} = {c},
  cell{5}{4} = {c},
  cell{5}{5} = {c},
  cell{5}{6} = {c},
  cell{5}{7} = {c},
  cell{5}{8} = {c},
  cell{5}{9} = {c},
  cell{5}{10} = {c},
  cell{5}{11} = {c},
  cell{5}{12} = {c},
  cell{5}{13} = {c},
  cell{5}{14} = {c},
  cell{5}{15} = {c},
  cell{5}{16} = {c},
  cell{6}{2} = {c},
  cell{6}{3} = {c},
  cell{6}{4} = {c},
  cell{6}{5} = {c},
  cell{6}{6} = {c},
  cell{6}{7} = {c},
  cell{6}{8} = {c},
  cell{6}{9} = {c},
  cell{6}{10} = {c},
  cell{6}{11} = {c},
  cell{6}{12} = {c},
  cell{6}{13} = {c},
  cell{6}{14} = {c},
  cell{6}{15} = {c},
  cell{6}{16} = {c},
  cell{7}{2} = {c},
  cell{7}{3} = {c},
  cell{7}{4} = {c},
  cell{7}{5} = {c},
  cell{7}{6} = {c},
  cell{7}{7} = {c},
  cell{7}{8} = {c},
  cell{7}{9} = {c},
  cell{7}{10} = {c},
  cell{7}{11} = {c},
  cell{7}{12} = {c},
  cell{7}{13} = {c},
  cell{7}{14} = {c},
  cell{7}{15} = {c},
  cell{7}{16} = {c},
  cell{8}{1} = {c=16}{0.916\linewidth},
  cell{9}{2} = {c},
  cell{9}{3} = {c},
  cell{9}{4} = {c},
  cell{9}{5} = {c},
  cell{9}{6} = {c},
  cell{9}{7} = {c},
  cell{9}{8} = {c},
  cell{9}{9} = {c},
  cell{9}{10} = {c},
  cell{9}{11} = {c},
  cell{9}{12} = {c},
  cell{9}{13} = {c},
  cell{9}{14} = {c},
  cell{9}{15} = {c},
  cell{9}{16} = {c},
  cell{10}{2} = {c},
  cell{10}{3} = {c},
  cell{10}{4} = {c},
  cell{10}{5} = {c},
  cell{10}{6} = {c},
  cell{10}{7} = {c},
  cell{10}{8} = {c},
  cell{10}{9} = {c},
  cell{10}{10} = {c},
  cell{10}{11} = {c},
  cell{10}{12} = {c},
  cell{10}{13} = {c},
  cell{10}{14} = {c},
  cell{10}{15} = {c},
  cell{10}{16} = {c},
  cell{11}{2} = {c},
  cell{11}{3} = {c},
  cell{11}{4} = {c},
  cell{11}{5} = {c},
  cell{11}{6} = {c},
  cell{11}{7} = {c},
  cell{11}{8} = {c},
  cell{11}{9} = {c},
  cell{11}{10} = {c},
  cell{11}{11} = {c},
  cell{11}{12} = {c},
  cell{11}{13} = {c},
  cell{11}{14} = {c},
  cell{11}{15} = {c},
  cell{11}{16} = {c},
  cell{12}{1} = {c=16}{0.916\linewidth},
  cell{13}{2} = {c},
  cell{13}{3} = {c},
  cell{13}{4} = {c},
  cell{13}{5} = {c},
  cell{13}{6} = {c},
  cell{13}{7} = {c},
  cell{13}{8} = {c},
  cell{13}{9} = {c},
  cell{13}{10} = {c},
  cell{13}{11} = {c},
  cell{13}{12} = {c},
  cell{13}{13} = {c},
  cell{13}{14} = {c},
  cell{13}{15} = {c},
  cell{13}{16} = {c},
  cell{14}{2} = {c},
  cell{14}{3} = {c},
  cell{14}{4} = {c},
  cell{14}{5} = {c},
  cell{14}{6} = {c},
  cell{14}{7} = {c},
  cell{14}{8} = {c},
  cell{14}{9} = {c},
  cell{14}{10} = {c},
  cell{14}{11} = {c},
  cell{14}{12} = {c},
  cell{14}{13} = {c},
  cell{14}{14} = {c},
  cell{14}{15} = {c},
  cell{14}{16} = {c},
  cell{15}{2} = {c},
  cell{15}{3} = {c},
  cell{15}{4} = {c},
  cell{15}{5} = {c},
  cell{15}{6} = {c},
  cell{15}{7} = {c},
  cell{15}{8} = {c},
  cell{15}{9} = {c},
  cell{15}{10} = {c},
  cell{15}{11} = {c},
  cell{15}{12} = {c},
  cell{15}{13} = {c},
  cell{15}{14} = {c},
  cell{15}{15} = {c},
  cell{15}{16} = {c},
  cell{16}{2} = {c},
  cell{16}{3} = {c},
  cell{16}{4} = {c},
  cell{16}{5} = {c},
  cell{16}{6} = {c},
  cell{16}{7} = {c},
  cell{16}{8} = {c},
  cell{16}{9} = {c},
  cell{16}{10} = {c},
  cell{16}{11} = {c},
  cell{16}{12} = {c},
  cell{16}{13} = {c},
  cell{16}{14} = {c},
  cell{16}{15} = {c},
  cell{16}{16} = {c},
  cell{17}{2} = {c},
  cell{17}{3} = {c},
  cell{17}{4} = {c},
  cell{17}{5} = {c},
  cell{17}{6} = {c},
  cell{17}{7} = {c},
  cell{17}{8} = {c},
  cell{17}{9} = {c},
  cell{17}{10} = {c},
  cell{17}{11} = {c},
  cell{17}{12} = {c},
  cell{17}{13} = {c},
  cell{17}{14} = {c},
  cell{17}{15} = {c},
  cell{17}{16} = {c},
  vline{2-3} = {1}{},
  vline{2,13} = {1-3,5-7,9-11,13-17}{},
  hline{1-5,8-9,12-13,18} = {-}{},
}
                                & Held In       &               &               &               &               &               &               &               &               &               &               & Held Out      &               &               &               \\
Dataset                         & WikiTQ        & HybridQA      & FeTaQA        & TabMWP        & WikiSQL       & TabFact       & ToTTo         & KVRet         & CWQ           & DART          & \textbf{Avg}  & SQA           & WTT           & FinQA         & \textbf{Avg}  \\
Metric                          & Ex            & Acc           & BLEU          & Acc           & Ex            & Acc           & BLEU          & Micro         & Acc           & BLEU          &               & Acc           & BLEU          & Acc           &               \\
\textbf{1-shot Learning}        &               &               &               &               &               &               &               &               &               &               &               &               &               &               &               \\
Mistral 7B Instruct             & 20.0          & 22.9          & 8.9           & 30.5          & 24.6          & 54.8          & 16.8          & 54.0          & 34.5          & 43.5          & 31.1          & 3.2           & 3.8           & 6.7           & 4.6           \\
ChatGPT 3.5                     & 42.6          & 38.4          & 15.1          & 52.9          & 50.4          & 53.0          & 22.4          & 53.0          & 50.1          & 57.0          & 43.5          & 9.7           & 4.0           & 12.2          & 8.6           \\
ChatGPT 4                       & 60.8          & 50.8          & 8.4           & 72.6          & 35.8          & 79.0          & 21.4          & 60.3          & 66.6          & 53.7          & 50.9          & 5.4           & 3.1           & 18.0          & 8.8           \\
\textbf{Full Parameters Tuning} &               &               &               &               &               &               &               &               &               &               &               &               &               &               &               \\
USKG 3B $\times$ N              & 49.3          & 59.2          & 36.0          & -             & 86.0          & 80.8          & 49.0          & 67.9          & 73.3          & 46.7          & -             & 0             & 0             & 0             & 0             \\
FLAN-UL2 20B                    & 54.6          & 61.0          & 35.8          & -             & 87.3          & \textbf{87.1} & -             & -             & 75.9          & 50.4          & -             & 70.1$\dag$          & 19.4$\dag$          & 5.9$\dag$           & 31.8$\dag$          \\
StructLM 7B-M                   & 56.8          & 62.6          & 37.5          & 73.5          & 87.0          & 84.6          & \textbf{49.8} & \textbf{72.2} & \textbf{79.9} & 63.2          & 66.7          & 41.9          & \textbf{16.7} & \textbf{24.6} & \textbf{27.7} \\
\textbf{Lora Tuning}            &               &               &               &               &               &               &               &               &               &               &               &               &               &               &               \\
TableLlama 7B                                                          & 35.0*         & 39.4*         & \textbf{39.0} & -             & 50.5*         & 82.5          & 20.8*         & 48.7*         & -             & -    & -             & 2.6           & 3.0           & 1.4           & 2.3                                   \\
Mistral 7B Instruct                                                    & 56.9          & 62.4          & 36.7          & \textbf{76.9} & 86.8          & 84.1          & 49.0          & 71.2          & 78.1          & 65.0 & 66.7          & 47.1          & 10.7          & 14.2          & 24.0                                  \\
HGT 7B-M                                                               & 57.2          & 62.4          & 36.6          & 75.8          & 87.0          & 83.8          & \textbf{49.2} & 70.8          & 78.3          & \textbf{65.4} & 66.7          & 51.0          & 8.2           & \textbf{18.0} & 25.7                                  \\
G-Retrieve 7B-M                                                        & \textbf{57.4} & 62.6          & 36.5          & 76.0          & 86.6          & 84.2          & 48.9          & 71.6          & 78.4          & 65.2 & 66.7          & 50.9          & 12.5          & 16.6          & 26.7                                  \\
LlaSA 7B-M (Ours)                                                      & 56.2          & \textbf{62.9} & 37.0          & 76.7          & \textbf{87.1} & \textbf{84.3} & 49.0          & \textbf{72.3} & \textbf{78.4} & 64.7 & \textbf{66.9} & \textbf{51.3} & \textbf{16.3} & 13.9          & \textbf{27.2}                         \\
\end{tblr}
\caption{The evaluation results of our model against other baselines. Cells with "*" represent that the model did not train on this dataset.  Cells in the held-out section with "$\dag$" are held-in results. 7B-M represents using Mistral-7B-Instruct-v0.2 as the base model. The results of HGT 7B-M and G-Retrieve 7B-M were re-implemented by  us. The results of StructLM 7B-M are from their paper. USKG 3B $\times$ N indicates training a 3B model for each task. The boldface indicates the best result.
}
\label{tab:main_results}
\end{table*}

\subsection{Training}

\subsubsection{Pretraining}
\label{sec:pretraining}

Even though HyTrel \cite{chen_hytrel_2023} also trains an encoder for table encoding, its pretraining tasks, such as column type classification and table similarity prediction, does not truly align the hypergraph encoder space with the textual space. Similar to Q-Former \cite{li_blip-2_2023}, we also introduce two tasks to effectively align these two spaces, and their attention mechanisms are shown in the upper left corner of Figure \ref{fig:method}. The details of constructing pretraining dataset can be founded in Appendix \ref{app:pretraining_dataset}.

\noindent \textbf{Graph-Depended
Answer Generation}. This tasks trains the G-Former to generate answers, given input tables as the condition. The information required for generating the answer is first extracted by the query tokens with cross-attention, and then passed to the text transformer through self-attention. The graph transformer learns to compress all the graph node representations into a fixed number of query tokens. The multimodal causal attention allow query tokens to interact with each others but not the text tokens while each text token can interact with all query tokens and its previous text tokens.

\noindent \textbf{Graph-Text
Matching}. Since some answers can be easily deduced even without any structural information, for example, the column name for "Miami Heat" is likely to be "Team". Therefore, we also introduce Graph-Text Matching. We employ a bi-directional self-attention mask, allowing all queries and text tokens to attend to each other. As a result, the output query embeddings, denoted as $Q$, integrate multimodal information effectively. Each query embedding is then passed through a MLP to generate a corresponding logit. Finally, we compute the overall matching score by averaging the logits across all queries.


\subsubsection{Task-specific Instruction Tuning}

In the instruction tuning stage, we use multiple SKG tasks to finetune LLMs through Parameter-Efficient Fine-Tuning \cite{ding2023peft}. In this stage, we only use the graph transformer module pretrained in the pretraining stage, that is, we extract fixed-length query embeddings $\hat{\mathbf{q}}$ from the node representations of hypergraph $\mathcal{G}$. Then the extracted query embeddings $\hat{\mathbf{q}}$ are projected into the same dimension as the text embedding of the LLM through a fully connected layer. This process is defined as follows:
\begin{align}
  \hat{\mathbf{q}} = \operatorname{FC}(f_g(\mathbf{q}, \mathbf{X}))
\end{align}
where $\mathbf{X} \in \mathcal{R}^{n \times d_1}$ is the hypergraph node embeddings, $n$ is the number of nodes in the graph, $d_1$ is the dimension of node embeddings, ${\mathbf{q}} \in \mathcal{R}^{b \times d_1}$ is the original query embeddings, 
$\hat{\mathbf{q}} \in \mathcal{R}^{m \times d_l}$ is the extracted query embeddings, $m$ is the number of query tokens, $d_l$ is the dimension of the LLM's text embeddings, $f_g$ and $FC$ represents G-Former and fully connected layer.

These projected query embeddings are treated as soft prompts and appended to the text embeddings. The LLM learns to predict answers based on these text embeddings and soft prompts. This process is defined as follows:
\begin{align}
\mathbf{h_t}=\operatorname{TextEmbedder}([\operatorname{serialize(\mathcal{G})};x_q]) \\
        p_\theta(Y|\mathcal{G}, x_q) = \prod_{i=1}^{r}p_\theta(y_i|y_{\leq i}, [\mathbf{h_t};\hat{\mathbf{q}}])
\end{align}
where $\theta$ is the LLM's parameters, $serialize$ denotes function that serializes structured data to text sequence, [$;$] represents concatenation operation, $x_q$ and $Y$ represents the question and answer, respectively.

\begin{table*}
\tiny
\centering
\begin{tblr}{
  rows={1.4em, m, rowsep=1pt},
  width = \linewidth,
  colspec = {Q[180]Q[58]Q[65]Q[60]Q[65]Q[63]Q[60]Q[54]Q[52]Q[50]Q[50]Q[46]Q[46]Q[52]Q[52]Q[46]},
  row{2} = {c},
  row{3} = {c},
  row{4} = {c},
  row{13} = {c},
  cell{1}{2} = {c=11}{0.623\linewidth,c},
  cell{1}{13} = {c=4}{0.196\linewidth,c},
  cell{4}{1} = {c=16}{0.917\linewidth},
  cell{5}{2} = {c},
  cell{5}{3} = {c},
  cell{5}{4} = {c},
  cell{5}{5} = {c},
  cell{5}{6} = {c},
  cell{5}{7} = {c},
  cell{5}{8} = {c},
  cell{5}{9} = {c},
  cell{5}{10} = {c},
  cell{5}{11} = {c},
  cell{5}{12} = {c},
  cell{5}{13} = {c},
  cell{5}{14} = {c},
  cell{5}{15} = {c},
  cell{5}{16} = {c},
  cell{6}{2} = {c},
  cell{6}{3} = {c},
  cell{6}{4} = {c},
  cell{6}{5} = {c},
  cell{6}{6} = {c},
  cell{6}{7} = {c},
  cell{6}{8} = {c},
  cell{6}{9} = {c},
  cell{6}{10} = {c},
  cell{6}{11} = {c},
  cell{6}{12} = {c},
  cell{6}{13} = {c},
  cell{6}{14} = {c},
  cell{6}{15} = {c},
  cell{6}{16} = {c},
  cell{7}{2} = {c},
  cell{7}{3} = {c},
  cell{7}{4} = {c},
  cell{7}{5} = {c},
  cell{7}{6} = {c},
  cell{7}{7} = {c},
  cell{7}{8} = {c},
  cell{7}{9} = {c},
  cell{7}{10} = {c},
  cell{7}{11} = {c},
  cell{7}{12} = {c},
  cell{7}{13} = {c},
  cell{7}{14} = {c},
  cell{7}{15} = {c},
  cell{7}{16} = {c},
  cell{8}{2} = {c},
  cell{8}{3} = {c},
  cell{8}{4} = {c},
  cell{8}{5} = {c},
  cell{8}{6} = {c},
  cell{8}{7} = {c},
  cell{8}{8} = {c},
  cell{8}{9} = {c},
  cell{8}{10} = {c},
  cell{8}{11} = {c},
  cell{8}{12} = {c},
  cell{8}{13} = {c},
  cell{8}{14} = {c},
  cell{8}{15} = {c},
  cell{8}{16} = {c},
  cell{9}{2} = {c},
  cell{9}{3} = {c},
  cell{9}{4} = {c},
  cell{9}{5} = {c},
  cell{9}{6} = {c},
  cell{9}{7} = {c},
  cell{9}{8} = {c},
  cell{9}{9} = {c},
  cell{9}{10} = {c},
  cell{9}{11} = {c},
  cell{9}{12} = {c},
  cell{9}{13} = {c},
  cell{9}{14} = {c},
  cell{9}{15} = {c},
  cell{9}{16} = {c},
  cell{10}{2} = {c},
  cell{10}{3} = {c},
  cell{10}{4} = {c},
  cell{10}{5} = {c},
  cell{10}{6} = {c},
  cell{10}{7} = {c},
  cell{10}{8} = {c},
  cell{10}{9} = {c},
  cell{10}{10} = {c},
  cell{10}{11} = {c},
  cell{10}{12} = {c},
  cell{10}{13} = {c},
  cell{10}{14} = {c},
  cell{10}{15} = {c},
  cell{10}{16} = {c},
  cell{11}{2} = {c},
  cell{11}{3} = {c},
  cell{11}{4} = {c},
  cell{11}{5} = {c},
  cell{11}{6} = {c},
  cell{11}{7} = {c},
  cell{11}{8} = {c},
  cell{11}{9} = {c},
  cell{11}{10} = {c},
  cell{11}{11} = {c},
  cell{11}{12} = {c},
  cell{11}{13} = {c},
  cell{11}{14} = {c},
  cell{11}{15} = {c},
  cell{11}{16} = {c},
  cell{12}{2} = {c},
  cell{12}{3} = {c},
  cell{12}{4} = {c},
  cell{12}{5} = {c},
  cell{12}{6} = {c},
  cell{12}{7} = {c},
  cell{12}{8} = {c},
  cell{12}{9} = {c},
  cell{12}{10} = {c},
  cell{12}{11} = {c},
  cell{12}{12} = {c},
  cell{12}{13} = {c},
  cell{12}{14} = {c},
  cell{12}{15} = {c},
  cell{12}{16} = {c},
  cell{13}{1} = {c=16}{0.917\linewidth},
  cell{14}{2} = {c},
  cell{14}{3} = {c},
  cell{14}{4} = {c},
  cell{14}{5} = {c},
  cell{14}{6} = {c},
  cell{14}{7} = {c},
  cell{14}{8} = {c},
  cell{14}{9} = {c},
  cell{14}{10} = {c},
  cell{14}{11} = {c},
  cell{14}{12} = {c},
  cell{14}{13} = {c},
  cell{14}{14} = {c},
  cell{14}{15} = {c},
  cell{14}{16} = {c},
  cell{15}{2} = {c},
  cell{15}{3} = {c},
  cell{15}{4} = {c},
  cell{15}{5} = {c},
  cell{15}{6} = {c},
  cell{15}{7} = {c},
  cell{15}{8} = {c},
  cell{15}{9} = {c},
  cell{15}{10} = {c},
  cell{15}{11} = {c},
  cell{15}{12} = {c},
  cell{15}{13} = {c},
  cell{15}{14} = {c},
  cell{15}{15} = {c},
  cell{15}{16} = {c},
  cell{16}{2} = {c},
  cell{16}{3} = {c},
  cell{16}{4} = {c},
  cell{16}{5} = {c},
  cell{16}{6} = {c},
  cell{16}{7} = {c},
  cell{16}{8} = {c},
  cell{16}{9} = {c},
  cell{16}{10} = {c},
  cell{16}{11} = {c},
  cell{16}{12} = {c},
  cell{16}{13} = {c},
  cell{16}{14} = {c},
  cell{16}{15} = {c},
  cell{16}{16} = {c},
  cell{17}{2} = {c},
  cell{17}{3} = {c},
  cell{17}{4} = {c},
  cell{17}{5} = {c},
  cell{17}{6} = {c},
  cell{17}{7} = {c},
  cell{17}{8} = {c},
  cell{17}{9} = {c},
  cell{17}{10} = {c},
  cell{17}{11} = {c},
  cell{17}{12} = {c},
  cell{17}{13} = {c},
  cell{17}{14} = {c},
  cell{17}{15} = {c},
  cell{17}{16} = {c},
  cell{18}{2} = {c},
  cell{18}{3} = {c},
  cell{18}{4} = {c},
  cell{18}{5} = {c},
  cell{18}{6} = {c},
  cell{18}{7} = {c},
  cell{18}{8} = {c},
  cell{18}{9} = {c},
  cell{18}{10} = {c},
  cell{18}{11} = {c},
  cell{18}{12} = {c},
  cell{18}{13} = {c},
  cell{18}{14} = {c},
  cell{18}{15} = {c},
  cell{18}{16} = {c},
  cell{19}{2} = {c},
  cell{19}{3} = {c},
  cell{19}{4} = {c},
  cell{19}{5} = {c},
  cell{19}{6} = {c},
  cell{19}{7} = {c},
  cell{19}{8} = {c},
  cell{19}{9} = {c},
  cell{19}{10} = {c},
  cell{19}{11} = {c},
  cell{19}{12} = {c},
  cell{19}{13} = {c},
  cell{19}{14} = {c},
  cell{19}{15} = {c},
  cell{19}{16} = {c},
  cell{20}{2} = {c},
  cell{20}{3} = {c},
  cell{20}{4} = {c},
  cell{20}{5} = {c},
  cell{20}{6} = {c},
  cell{20}{7} = {c},
  cell{20}{8} = {c},
  cell{20}{9} = {c},
  cell{20}{10} = {c},
  cell{20}{11} = {c},
  cell{20}{12} = {c},
  cell{20}{13} = {c},
  cell{20}{14} = {c},
  cell{20}{15} = {c},
  cell{20}{16} = {c},
  cell{21}{2} = {c},
  cell{21}{3} = {c},
  cell{21}{4} = {c},
  cell{21}{5} = {c},
  cell{21}{6} = {c},
  cell{21}{7} = {c},
  cell{21}{8} = {c},
  cell{21}{9} = {c},
  cell{21}{10} = {c},
  cell{21}{11} = {c},
  cell{21}{12} = {c},
  cell{21}{13} = {c},
  cell{21}{14} = {c},
  cell{21}{15} = {c},
  cell{21}{16} = {c},
  vline{2-3} = {1}{},
  vline{2,13} = {1-3,5-12,14-21}{},
  hline{1-5,7,9,11,13-14,16,18,20,22} = {-}{},
}
                             & Held In       &               &               &               &               &               &               &               &               &               &               & Held Out      &               &               &               \\
dataset                      & WikiTQ        & HybridQA      & FeTaQA        & TabMWP        & WikiSQL       & TabFact       & ToTTo         & KVRet         & CWQ           & DART          & \textbf{Avg}  & SQA           & WTT           & FinQA         & \textbf{Avg}  \\
metric                       & Ex            & Acc           & BLEU          & Acc           & Ex            & Acc           & BLEU          & Micro         & Acc           & BLEU          &               & Acc           & BLEU          & Acc           &               \\
\textbf{\textbf{Freeze LLM}} &               &               &               &               &               &               &               &               &               &               &               &               &               &               &               \\
Phi 3B                       & 31.2          & 39.0          & 13.6          & 41.9          & 47.1          & 59.9          & 32.4          & 46.3          & 45.1          & 58.0          & 41.5          & 14.8          & 8.1           & 9.2           & 10.7          \\
LlaSA-Phi 3B                 & \textbf{35.1} & \textbf{49.1} & \textbf{26.6} & \textbf{62.5} & \textbf{65.4} & \textbf{70.3} & \textbf{39.9} & \textbf{62.1} & \textbf{60.2} & \textbf{59.7} & \textbf{53.1} & \textbf{21.8} & \textbf{11.7} & \textbf{13.7} & \textbf{15.7} \\
Llama2 7B                    & 27.2          & 42.3          & 8.9           & 27.3          & 45.0          & 50.8          & 32.7          & 46.8          & 52.2          & 51.6          & 38.5          & 11.0          & 7.7           & 2.1           & 6.9           \\
LlaSA-Llama2 7B              & \textbf{32.5} & \textbf{50.2} & \textbf{26.6} & \textbf{48.5} & \textbf{62.9} & \textbf{66.2} & \textbf{40.6} & \textbf{62.0} & \textbf{62.1} & \textbf{58.3} & \textbf{51.0} & \textbf{19.9} & \textbf{17.5} & \textbf{3.5}  & \textbf{13.6} \\
Mistral 7B                   & 34.1          & 44.2          & 5.3           & 37.9          & 55.3          & 59.4          & 35.2          & 43.7          & 55.4          & 58.1          & 42.9          & \textbf{26.7} & \textbf{14.8} & \textbf{12.1} & \textbf{17.9} \\
LlaSA-Mistral 7B             & \textbf{38.4} & \textbf{50.6} & \textbf{27.3} & \textbf{59.9} & \textbf{70.1} & \textbf{73.6} & \textbf{42.4} & \textbf{65.7} & \textbf{67.1} & \textbf{59.2} & \textbf{55.4} & 25.9          & 6.0           & 7.4           & 13.1          \\
Llama3 8B                    & 41.2          & 50.1          & 20.3          & 52.7          & 67.0          & 66.1          & 38.2          & 53.2          & 62.4          & 59.3          & 51.1          & \textbf{31.2} & \textbf{13.0} & 17.0          & 20.4          \\
LlaSA-Llama3 8B              & \textbf{45.9} & \textbf{53.8} & \textbf{29.9} & \textbf{70.5} & \textbf{74.8} & \textbf{78.2} & \textbf{43.1} & \textbf{64.3} & \textbf{69.6} & \textbf{60.4} & \textbf{59.1} & 29.0          & 12.2          & \textbf{23.1} & \textbf{21.4} \\
\textbf{Lora Tuning LLM}     &               &               &               &               &               &               &               &               &               &               &               &               &               &               &               \\
Phi 3B                       & 45.8          & 53.6          & 30.7          & 70.0          & 80.2          & 75.5          & 42.6          & 62.5          & 68.3          & 62.9          & 59.2          & 34.3          & 9.5           & \textbf{11.4}          & 18.4          \\
LlaSA-Phi 3B                 & \textbf{47.4} & \textbf{55.4} & \textbf{31.6} & \textbf{72.4} & \textbf{81.5} & \textbf{77.5} & \textbf{44.5} & \textbf{67.8} & \textbf{70.8} & 62.0          & \textbf{61.1} & \textbf{45.8} & \textbf{13.0} & 7.1  & \textbf{22.0} \\
Llama2 7B                    & 45.0          & 59.5          & 32.5          & 62.8          & 82.9          & 78.1          & 46.3          & \textbf{67.1} & 75.0          & 63.8          & 61.3          & 35.3          & 8.6           & 6.5           & 16.8          \\
LlaSA-Llama2 7B              & \textbf{45.9} & \textbf{60.0} & \textbf{33.0} & \textbf{64.2} & \textbf{83.0} & \textbf{78.6} & \textbf{47.0} & 66.6          & \textbf{76.0} & \textbf{63.1} & \textbf{61.7} & \textbf{35.5} & \textbf{8.6} & \textbf{9.7} & \textbf{17.9} \\
Mistral 7B                                                    & \textbf{56.9}          & 62.4          & 36.7          & \textbf{76.9} & 86.8          & 84.1          & 49.0          & 71.2          & 78.1          & \textbf{65.0} & 66.7          & 47.1          & 10.7          & \textbf{14.2}          & 24.0                                  \\
LlaSA-Mistral 7B                & 56.2          & \textbf{62.9} & \textbf{37.0}          & 76.7          & \textbf{87.1} & \textbf{84.3} & \textbf{49.0}          & \textbf{72.3} & \textbf{78.4} & 64.7 & \textbf{66.9} & \textbf{51.3} & \textbf{16.3} & 13.9          & \textbf{27.2}   \\
Llama3 8B                    & 59.4          & 62.8          & 34.1          & 77.0          & 86.2          & 85.8          & 47.7          & 69.2          & 78.9          & \textbf{64.0} & 66.5          & 47.7          & \textbf{11.7} & 21.5          & 27.0          \\
LlaSA-Llama3 8B              & \textbf{60.4} & \textbf{63.0} & \textbf{34.7} & \textbf{77.8} & \textbf{86.3} & \textbf{86.1} & \textbf{48.0} & \textbf{69.2} & \textbf{79.0} & 63.1          & \textbf{66.8} & \textbf{52.6} & 10.8          & \textbf{22.8} & \textbf{28.7} 
\end{tblr}
\caption{The evaluation results of LLaSA with different base models under different finetuning strategies. The soft tokens in prompt tuning (Freeze LLM) is set 10, which is the same as the number of query tokens in G-Former. The lora rank is set to 32 in lora tuning.}
\label{tab:different_base_models}
\end{table*}
\section{Experiment}

\subsection{Datasets}

To validate the effectiveness of our approach, we collected 10 SKG tasks as our training data, which can be categorized into the following four types: 
(1) \textbf{Structured Data Question Answering}: This task requires the LLM to answer questions based on the given tables, knowledge graphs, and textual information. The datasets for this category include WikiTQ \cite{pasupat2015wtq}, CompWebQ \cite{talmor2018cwq}, and TabMWP \cite{lu2022tabmwp}. 
(2) \textbf{Fact Verification}: This task requires the LLM to determine whether a given statement is \textit{entailed} or \textit{refuted} based on the information in the table. The corresponding dataset is TabFact \cite{chen2019tabfact}. 
(3) \textbf{Structured Data to Text}: This task requires the LLM to summarize or describe the content of a given table or knowledge graph in one or two sentences. The relevant datasets for this category include ToTTo \cite{parikh2020totto} and DART \cite{nan2020dart}.

To evaluate the generalization ability of our method, we use SQA\cite{iyyer2017sqa}, WikiTableText \cite{bao2018wikitabletext} and FinQA \cite{chen2021finqa} as held-out datasets, where SQA belongs to table-based question, WikiTableText belongs to structured data to Text and FinQA requires generating python-executable math expression based on the given questions and tables.

Statistics of these datasets can be found in Appendix \ref{app:skg_datasets}.

\subsection{Baselines}
In this work, we compare LLaSA with other LLMs based methods. 
We primarily select StructLM \cite{zhuang_structlm_2024}, which performs full parameters fine-tuning on various SKG datasets, as the main baseline. It is important to note that StructLM utilizes a broader range of datasets such as SQL2Text \cite{shu2021sql2text}. These datasets are excluded from LLaSA's training set because their inputs could not be transformed into hypergraphs. We also compare with TableLLama \cite{zhang_tablellama_2024}, which not only leverages a broader range of foundational table tasks, such as Column Type Annotation and Entity Linking, but also uses a longer 8K context length to finetune the LLMs. As HGT\cite{jin_hgt_2024} and G-Retrieve\cite{he_g-retriever_2024} use different models and training datasets, we re-implement and train them under our framework for a fairer comparison, where HGT concatenates the representations of all GNN nodes to the LLM input, G-Retrieve takes the average of all GNN node representations and concatenate it to the LLM input.
Additionally, we also evaluate the performance of GPT-3.5, GPT-4, and Mistral-7B-Instruct-v0.2 \cite{jiang2023mistral7b} under a 1-shot setting. 

\subsection{Implement Details}

We choose Phi-3B \cite{abdin2024phi3technicalreporthighly}, LLama2-7B \cite{touvron2023llama2openfoundation}, Mistral-7B \cite{jiang2023mistral7b} and LLama3-8B \cite{dubey2024llama3herdmodels} as our base models. We use a learning rate of 2e-5 with a 3\% warm-up cosine scheduler, set the batch size as 3, epoch as 3. The default lora rank is set to 32. All this models are trained on 8 H800 80G using DeepSpeed ZeRO-2 \cite{aminabadi2022deepspeedinferenceenablingefficient}. A training on a 7B model takes about 12 hours. The maximum sequence length is set to 2048 during training, the maximum generation length to set as 1024 during inference. We set the dimension of the hypergraph encoder to 768, with 12 layers, and use RoBERTa-base as the initial parameters for the G-Former. The total number of parameters for both components is 400M. The G-Former and GNN are pretrained on 25M tables for one epoch, and these data are taken from TaBERT \cite{yin_tabert_2020}.

\subsection{Main results}
Table \ref{tab:main_results} presents the results of our LLaSA compared to previous baselines across 10 datasets.
From the table, we can see that GPT-3.5 and GPT-4 still fall short in handling SKG tasks, trailing behind LLaSA 7B-M by 23.3\% and 15.9\% points, respectively, across the ten tasks. Moreover, our LLaSA 7B-M achieves state-of-the-art (SOTA) performance in 4 out of 10 tasks within the LLM-based method.

It can be found that  HGT 7B-M and G-Retrieve 7B-M did not achieve much improvement compared to the naive LLM, which may be due to the following reasons: 
1) \textbf{The projector-based strategy introduces noise by feeding all node representations into the LLM.} In real-world structured data question-answering scenarios, many cells are irrelevant to the current question and may distract the LLM. In contrast, our G-Former compresses all graph node representations into a fixed-length token sequence, retaining only the most relevant information.
2) \textbf{The projector-based strategy fails to fully leverage the alignment objectives in pretraining.} During pretraining, we aligned the Q-Former, which acts as a bridge between GNN representations and the textual space, rather than directly aligning the GNN itself. Consequently, even pretrained GNNs cannot directly enhance the final performance of the LLM.

From the perspective of model parameters, we find that the performance of LoRA-tuned Mistral 7B closely approaches that of the fully fine-tuned StructLM 7B-M. When using LLaSA framework, the LoRA-tuned Mistral 7B can surpass StructLM 7B-M, despite the former only requiring 400M trainable parameters compared to the latter's 7B parameters, and outperforms StructLM 7B-M on 6 tasks.
Additionally, we observe that fully fine-tuning LLMs on SKG tasks may lead to a decline in their performance on other tasks, whereas LoRA-tuned LLMs experience a smaller drop. One piece of evidence is the TabMWP dataset, which requires mathematical reasoning, where LLaSA 7B-M significantly outperforms StructLM 7B-M by 4.2\%.

On the held-out data, although StructLM 7B achieves a higher average performance, our LLaSA significantly outperforms StructLM 7B on the SQA dataset.

\subsection{LLaSA with Different Base Models}

To verify the generality of our pretrained hypergraph encoder and G-Former, we evaluate LLaSA under different base models with different finetuning strategies (Prompt Tuning and Lora Tuning), the results are demonstrated in Table \ref{tab:different_base_models}.

As shown in the table, the pretrained hypergraph encoder and G-Former enhance the models' ability to handle SKG tasks and improve their generalization to unseen datasets across most base models (except for the prompt-tuned LLaSA-Mistral 7B, which shows a performance drop on the held-out data). Especially under the Freeze LLM setting, LLaSA achieves significant improvements compared to basic prompt tuning. Specifically, it delivers an approximate 10\% performance boost across Phi-3B, Llama2-7B, Mistral-7B, and Llama3-8B models. \textbf{This indicates that our pre-trained hypergraph encoder and G-Former can be effectively adapted to various LLMs, enhancing their ability to handle structured data. } 

Under the LoRA tuning LLM setting, although LLaSA achieves smaller improvements on the held-in datasets, it consistently enhances model performance on held-out datasets. This suggests that our approach genuinely improves the model's ability of handling structured data rather than merely overfitting to the training data. Additionally, we observed that LLaSA's 0.9\% performance improvement on Phi-3B is notably greater than the 0.3\% improvement on Llama3-8B. This difference may be attributed to Phi-3B's inherently weaker ability to process structured data, making the introduction of the hypergraph encoder more impactful in enhancing its performance.

\subsection{Comparison Between Different Pretraining Strategies}

\begin{figure}[t]
  \centering
  \includegraphics[width=0.48\textwidth]{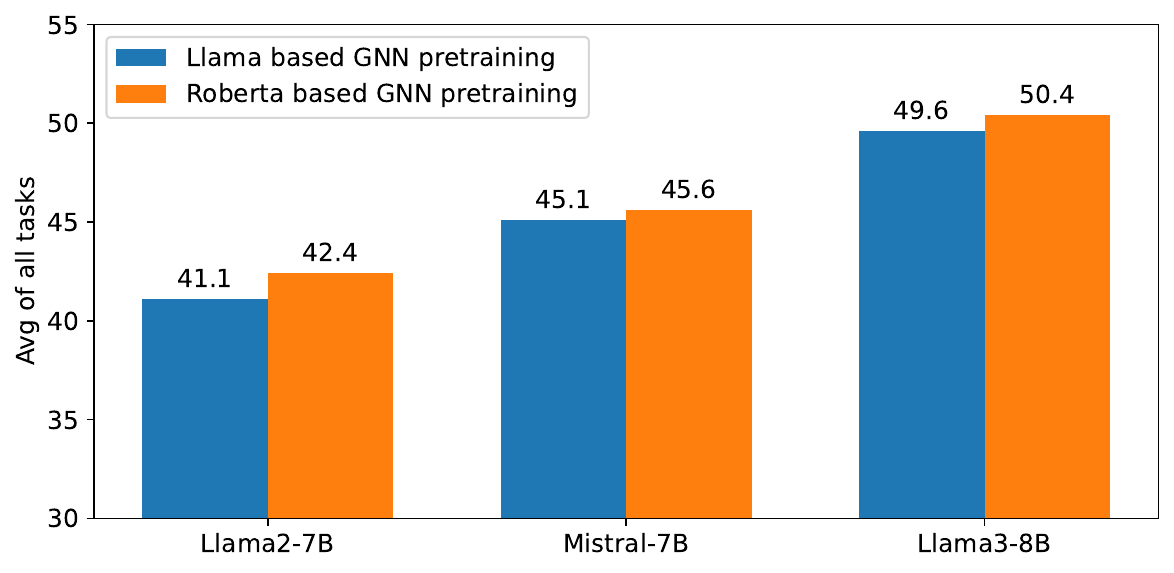} 
  \caption{The average performance of models using different pretrained Hypergraph Encoder (GNN).
  }
  \label{fig:pretraining_strategy_res}
\end{figure}

We compared two strategies for pre-training the GNN:
(1) Llama based pretraining, which use question answering task to pretrain a GNN based on a frozen LLM, as shown in Figure \ref{fig:pretraining_strategies} (a).
(2) G-Former-based pretraining, using both answer prediction and Graph-Text Matching tasks, as shown in Figure \ref{fig:pretraining_strategies} (b).
We select checkpoints with the same training time for both strategies and conduct experiments under the frozen LLM setting, the results are shown in Figure \ref{fig:pretraining_strategy_res}. The experimental results indicate that the GNN pretrained with G-Former exhibits superior adaptability, showing greater improvements across various models compared to the GNN pretrained with Llama. 
The main reasons are as follows: (1) In the Llama based pretraining, the soft tokens are not essential since serialized text is already included in the input, which prevents the GNN from fully aligning with the textual embedding space. As a result, it is unclear whether the GNN effectively encodes the table data as expected or just helps the LLMs fit the training data better. (2) The GNN pretrained with Llama primarily aligns with the Llama text space, limiting its adaptability to other models.

\subsection{Ablation Study}

\begin{table}
\small
\centering
\begin{tblr}{
  width = \linewidth,
  colspec = {Q[500]Q[210]Q[210]Q[210]},
  row{1} = {c},
  cell{2}{2} = {c},
  cell{2}{3} = {c},
  cell{2}{4} = {c},
  cell{3}{2} = {c},
  cell{3}{3} = {c},
  cell{3}{4} = {c},
  cell{4}{2} = {c},
  cell{4}{3} = {c},
  cell{4}{4} = {c},
  cell{5}{2} = {c},
  cell{5}{3} = {c},
  cell{5}{4} = {c},
  cell{6}{2} = {c},
  cell{6}{3} = {c},
  cell{6}{4} = {c},
  hline{1-3,7} = {-}{},
}
Method           & Avg-I       & Avg-O        & Avg           \\
LLaSA Llama-7B   & \textbf{51.0} & \textbf{13.6} & \textbf{32.3} \\
w/o pretraining  & 47.2          & 8.6           & 27.9          \\
w/o GNN          & 42.2          & 7.3           & 24.8          \\
w/o G-Former & 42.7          & 7.7           & 25.2        \\  
prompt tuning & 38.5          & 6.9           & 22.7        \\  
\end{tblr}
\caption{Ablation results on LLaSA Llama-7B. Avg-I and Avg-O represent the average score of Held-In and Held-Out datasets. 
The "w/o pretraining": randomly initializing GNN and G-Former without pre-training. 
The "w/o GNN": ignoring cross attention in G-Former.
The "w/o G-Former": ignoring the whole G-Former.
}
\label{tab:ablation}
\end{table}

Table \ref{tab:ablation} presents the results of our ablation study on LLaSA Llama2-7B under the frozen LLM setting. From the table, we can observe that compared to the randomly initialized GNN, the pretrained GNN helps the LLM achieve improvements of 3.8\% on Held-In datasets and 5.0\% on Held-Out datasets. This clearly demonstrates the effectiveness of the pretraining process.

In the "w/o GNN" and "w/o G-Former" settings, hypergraph information is ignored. The former directly passes the query token through multiple layers of self-attention, while the latter only applies a linear transformation via a fully connected layer. They can be viewed as more complex forms of prompt tuning. Although these two settings achieved a small 4\% improvement on Held-In datasets compared to basic prompt tuning, they do not show significant gains on Held-Out datasets. This suggests that simple prompt tuning mainly helps the model fit the training data better, without truly enhancing its generalization capability.

\section{Conclusion}

In this work, we propose LLaSA, a framework that converts structured data into hypergraphs and integrates the hypergraphs representations as an additional modality into the input of LLMs. We pretrain the hypergraph encoder on 25M tables with self-supervised learning. The experimental results on different LLMs over multiple datasets demonstrate the effectiveness and generalization of our method.

\section*{Limitation}
The limitations of our proposed LLaSA are as follows:
(1) We used a fixed number of query tokens, but the number of nodes in the hypergraph varies significantly, with some graphs having as few as a dozen nodes and others having over a hundred. As a result, when faced with graphs that have a large number of nodes, the G-Former may struggle to capture information effectively.
(2) Due to resource constraints, we conduct our experiments using a context length of 2K instead of the 8K used in TableLlama. The performance of LLaSA in longer contexts remains to be evaluated further.

\section*{Ethics Statement}
This paper proposes a method for SKG, and the experiments are conducted on public available datasets. 
As a result, there is no data privacy concern. Meanwhile, this paper does not involve human annotations, and there are no related ethical concerns.

\section*{Acknowledgment}
This work was supported by the National Key R\&D Program of China (No. 2022ZD0160503) and Beijing Natural Science Foundation (L243006) and the National Natural Science Foundation of China (No.62376270).

\bibliography{custom}
\begin{table*}
\scriptsize
\centering
\begin{tblr}{
  width = \linewidth,
  colspec = {Q[112]Q[60]Q[73]Q[63]Q[62]Q[71]Q[77]Q[67]Q[60]Q[62]Q[71]Q[77]Q[67]},
  cells = {c},
  row{1-3} = {m},
  cell{1}{2} = {c=2}{0.133\linewidth},
  cell{1}{4} = {c=5}{0.34\linewidth},
  cell{1}{9} = {c=5}{0.337\linewidth},
  vline{2-3,5} = {1}{},
  vline{2,4,9} = {1-15}{},
  hline{1-3,13,16} = {-}{},
}
              & Overall Length &                 & Train  &                 &                 &                   &          & Test  &                &                 &                   &          \\
Dataset       & {Input\\(avg)} & {Output\\(avg)} & Count  & {Input \\(max)} & {Output\\(max)} & {\# Nodes\\(avg)} & \# Trunc & Count & {Input\\(max)} & {Output\\(max)} & {\# Nodes\\(avg)} & \# Trunc \\
TabMWP        & 208            & 5               & 23059  & 709             & 33              & 20                & 0        & 7686  & 703            & 31              & 19                & 0        \\
ToTTo         & 252            & 31              & 120761 & 2040            & 155             & 110               & 467      & 7700  & 2048           & 119             & 111               & 31       \\
KVRet         & 573            & 17              & 6288   & 1217            & 161             & 57                & 0        & 807   & 1147           & 82              & 56                & 0        \\
HybridQA      & 700            & 7               & 62682  & 2047            & 91              & 92                & 200      & 3466  & 2048           & 79              & 93                & 6        \\
CompWebQ      & 1350           & 12              & 27639  & 2047            & 321             & 265               & 321      & 2816  & 2048           & 256             & 264               & 8        \\
TabFact       & 660            & 5               & 92283  & 2045            & 5               & 94                & 2        & 12779 & 1687           & 4               & 93                & 0        \\
WikiTQ        & 832            & 6               & 11321  & 2028            & 273             & 114               & 0        & 4344  & 2048           & 148             & 115               & 10       \\
WikiSQL       & 689            & 7               & 56355  & 2047            & 518             & 96                & 16       & 15878 & 2048           & 244             & 98                & 1        \\
FeTaQA        & 653            & 39              & 7326   & 1853            & 158             & 97                & 0        & 2003  & 1548           & 114             & 95                & 0        \\
DART          & 134            & 30              & 62659  & 406             & 258             & 17                & 0        & 5097  & 261            & 109             & 17                & 0        \\
SQA           & 657            & 35              & 12275  & 1812            & 1012            & 98                & 2        & 3011  & 1725           & 769             & 102               & 0        \\
WikiTableText & 150            & 27              & 10000  & 313             & 97              & 13                & 0        & 2000  & 226            & 89              & 14                & 0        \\
Finqa         & 1230           & 21              & 6251   & 2040            & 72              & 29                & 186      & 1147  & 2048           & 61              & 31                & 25       
\end{tblr}
\caption{The statistics of numbers of input and output tokens in the training and test sets for each task. "\# Trunc" indicates the number of samples where the input length exceeds 2048 tokens and has been truncated. "\# Nodes" indicates the average number of nodes in hypergraphs.}
\label{tab:dataset_stats}

\end{table*}
\newpage

\appendix

\section{Pretraining Dataset}
\label{app:pretraining_dataset}

We use the 25 million tabled collected by TaBERT \cite{yin_tabert_2020}. We designed three types of question templates and used them to generate 10 questions for each table. The specific templates are as follows: 
\begin{enumerate}
    \item \textit{What's the column name of "\{node\_name\}"
    ?}
    \item \textit{In the row where the value of \{first\_col\_name\} is "\{row\_value\}", what is the corresponding value of \{col\_name\}?}
    \item \textit{Are "\{node\_name1\}" and "\{node\_name2\}" in the same row?}.
\end{enumerate}

\section{SKG Datasets}
\label{app:skg_datasets}
Some datasets are not used in our study, such as FEVEROUS and Infotabs, because the tables in these datasets are not well-structured, with some rows having a different number of cells than the table headers.
The statistics of the SKG datasets we used are shown in Table \ref{tab:dataset_stats}. 

\end{document}